\documentclass[10pt,twocolumn,letterpaper]{article}

\usepackage[pagenumbers]{iccv} 

\definecolor{iccvblue}{rgb}{0.21,0.49,0.74}
\usepackage[pagebackref,breaklinks,colorlinks,allcolors=iccvblue]{hyperref}

\usepackage{footmisc}

\def\paperID{8594} 
\def\confName{ICCV}
\def\confYear{2025}

\title{DicFace: Dirichlet-Constrained Variational Codebook Learning for Temporally Coherent Video Face Restoration}

\author{%
  Yan Chen$^{1*}$,
  Hanlin Shang$^{1*}$,
  Ce Liu$^{1*}$,
  Yuxuan Chen$^{1}$,
  Hui Li$^{1}$,
  Weihao Yuan$^{2}$,\\
  Hao Zhu$^{3}$,
  Zilong Dong$^{2}$,
  Siyu Zhu$^{1\dagger}$ \\
  $^1$Fudan University \ \ \   
  $^2$Alibaba Group \ \ \   
  $^3$Nanjing University
}

\begin{document}
\maketitle
\begin{abstract}
Video face restoration faces a critical challenge in maintaining temporal consistency while recovering fine facial details from degraded inputs. 
This paper presents a novel approach that extends Vector-Quantized Variational Autoencoders (VQ-VAEs), pretrained on static high-quality portraits, into a video restoration framework through variational latent space modeling. 
Our key innovation lies in reformulating discrete codebook representations as Dirichlet-distributed continuous variables, enabling probabilistic transitions between facial features across frames. 
A spatio-temporal Transformer architecture jointly models inter-frame dependencies and predicts latent distributions, while a Laplacian-constrained reconstruction loss combined with perceptual (LPIPS) regularization enhances both pixel accuracy and visual quality. 
Comprehensive evaluations on blind face restoration, video inpainting, and facial colorization tasks demonstrate state-of-the-art performance.
This work establishes an effective paradigm for adapting intensive image priors, pretrained on high-quality images,  to video restoration while addressing the critical challenge of flicker artifacts. The source code has been open-sourced and is available at \url{https://github.com/fudan-generative-vision/DicFace}.
\end{abstract}

\footnote{$^*$ These authors contribute equally to this work.}
    

\begin{figure*}[htbp]
    \centering
    \includegraphics[width=1.0\textwidth]{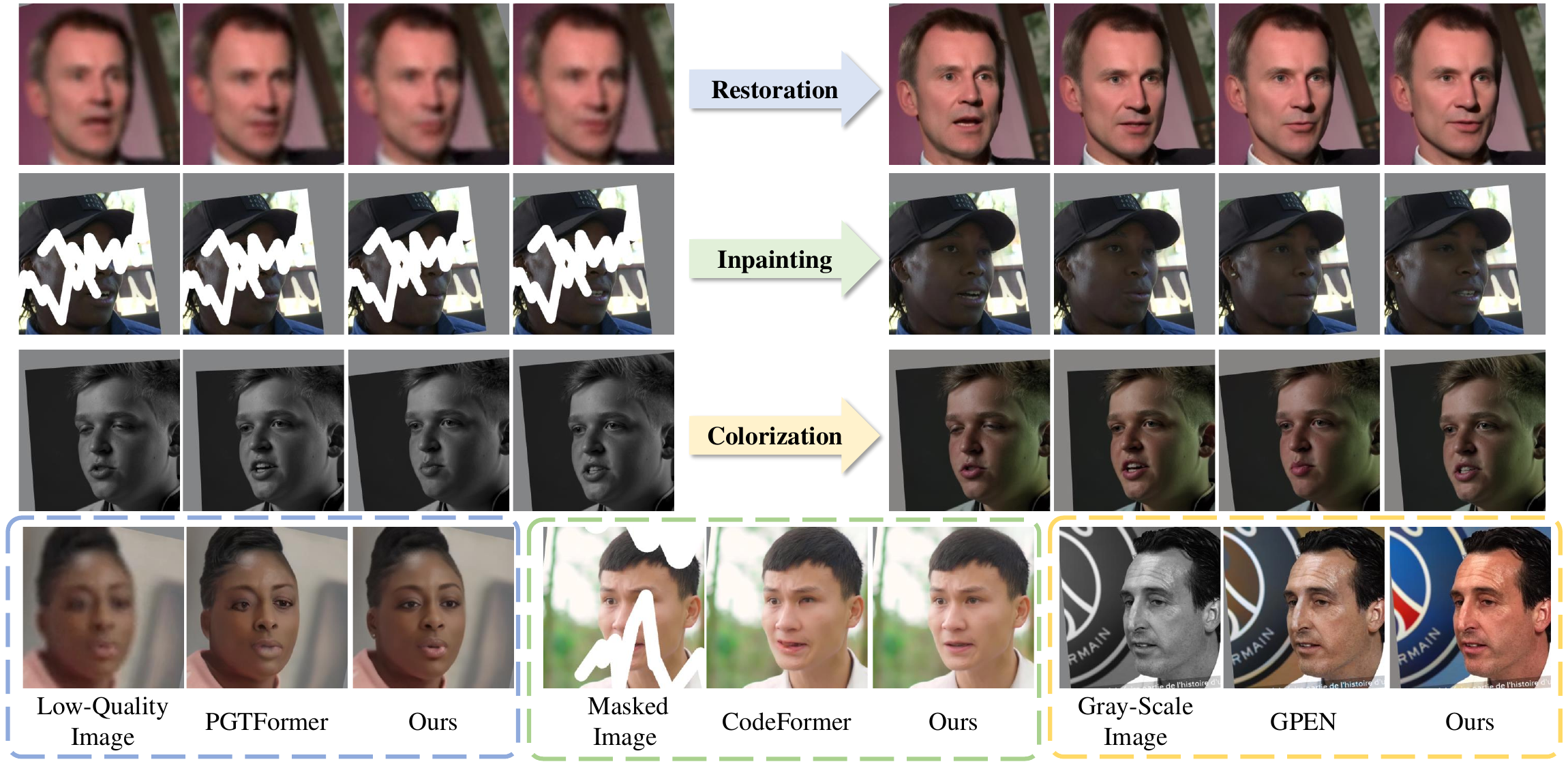}

    \vspace{-3mm}
    \caption{Our innovative facial restoration model exhibits markedly superior performance compared to existing state-of-the-art single-task methods (e.g. PGTFormer~\cite{xu2024beyond} for restoration ,CodeFormer~\cite{zhou2022towards} for inpainting and GPEN~\cite{yang2021gan} for colorization). We have successfully developed a more precise and natural restoration model that effectively maintains the facial structure and detailed characteristics of subjects. Furthermore, our approach enhances color consistency and ensures temporal continuity. This advancement represents a significant breakthrough in the domain of high-quality facial video and portrait restoration.}
    \label{fig:teaser}
    \vspace{-3mm}
\end{figure*}

\section{Introduction}\label{sec:intro}
Video face restoration aims to reconstruct high-quality face video sequences from degraded inputs. 
As a significant research branch in the field of computer vision—particularly within low-level vision and face analysis—it has important applications in digital image enhancement, film and media post-production, identity recognition, and security. 
The video face restoration problem is a domain-specific video restoration task and remains highly ill-posed, necessitating auxiliary guidance—such as facial codebooks~\cite{gu2022vqfr, zhou2022towards}, geometric priors~\cite{chen2018fsrnet, yu2018face}, and reference priors~\cite{li2018learning, li2020blind}—to enhance fidelity and improve visual details of reconstructed facial images.
Meanwhile, image-based face restoration~\cite{yang2021gan, wang2021towards, zhou2022towards} has benefited substantially from advances in visual generative models, which enable both the mapping from degraded low-quality inputs to high-quality outputs and the supplementation of fine details that are absent in the degraded inputs.

One straightforward approach ~\cite{wang2021towards, yang2021gan, zhou2022towards} for video face restoration is to directly apply image-based face restoration techniques to each frame of a low-quality input video, producing a sequence of reconstructed high-quality frames. 
However, simply extending image-based methods~\cite{yang2021gan, wang2021towards, zhou2022towards} to videos often fails to preserve the temporal consistency of reconstructed facial details, as each frame is typically processed with independently defined facial priors or code predictions.
An alternative commonly employed strategy is to adopt general video restoration methods~\cite{li2019fast,wang2019edvr,isobe2020video,chan2021basicvsr,chan2022basicvsr++} and adapt them to the video face restoration domain~\cite{feng2024kalman,xu2024beyond} through data-driven fine-tuning or domain adaptation. 
Nonetheless, these general video restoration methods often lack crucial facial priors (e.g., facial codebooks, geometric constraints, and reference priors), resulting in insufficiently detailed facial features and diminished visual fidelity in the restored output.

In this paper, we extend the classical vector-quantized autoencoder (VQ-AE)–based face restoration paradigm~\cite{esser2021taming}, which is pretrained on large-scale high-fidelity portrait images to form a high-quality codebook prior, from single-frame image restoration to multi-frame video restoration. 
Traditional methods~\cite{esser2021taming, gu2022vqfr, zhou2022towards} rely on a Transformer to independently predict discrete codes for each frame via codebook lookup, yet such per-frame processing disregards temporal continuity. 
Consequently, the latent codes may fluctuate abruptly across adjacent frames, causing perceptual flicker and compromising visual quality.
To address this limitation, we propose a novel variational formulation that bridges discrete and continuous representations. 
Specifically, we relax the discrete codebook by treating convex combinations of its items as Dirichlet-distributed latent variables. 
A Transformer then learns spatial-temporal dependencies over consecutive frames, while a Laplacian assumption on the reconstruction error together with LPIPS loss~\cite{zhang2018perceptual} promotes perceptually compelling results. 
By embedding discrete codebook representations into this continuous framework, the proposed strategy effectively mitigates temporal flicker in multi-frame outputs and achieves high-quality reconstructions through a principled variational training scheme.


Extensive experiments on the VFHQ~\cite{xie2022vfhq} benchmark demonstrate our method's effectiveness: quantitative evaluations on blind video face restoration task reveal that, compared to baseline methods, our approach achieves 1.27dB PSNR and 8.2\% LPIPS~\cite{zhang2018perceptual} improvements for image quality. Additionally, in terms of temporal stability, our method shows 5.6\% improvement in TLME metric. While qualitative assessments reveal enhanced temporal consistency under challenging conditions like occlusions and rapid motions. 
Furthermore, ablation studies validate our design choices, particularly the importance of Dirichlet-based variational modeling for temporal coherence. 
These advancements establish new state-of-the-art performance while providing a promising framework for integrating discrete facial priors with continuous video dynamics.

\begin{figure*}[htbp]
    \centering
    \includegraphics[width=1.0\textwidth]{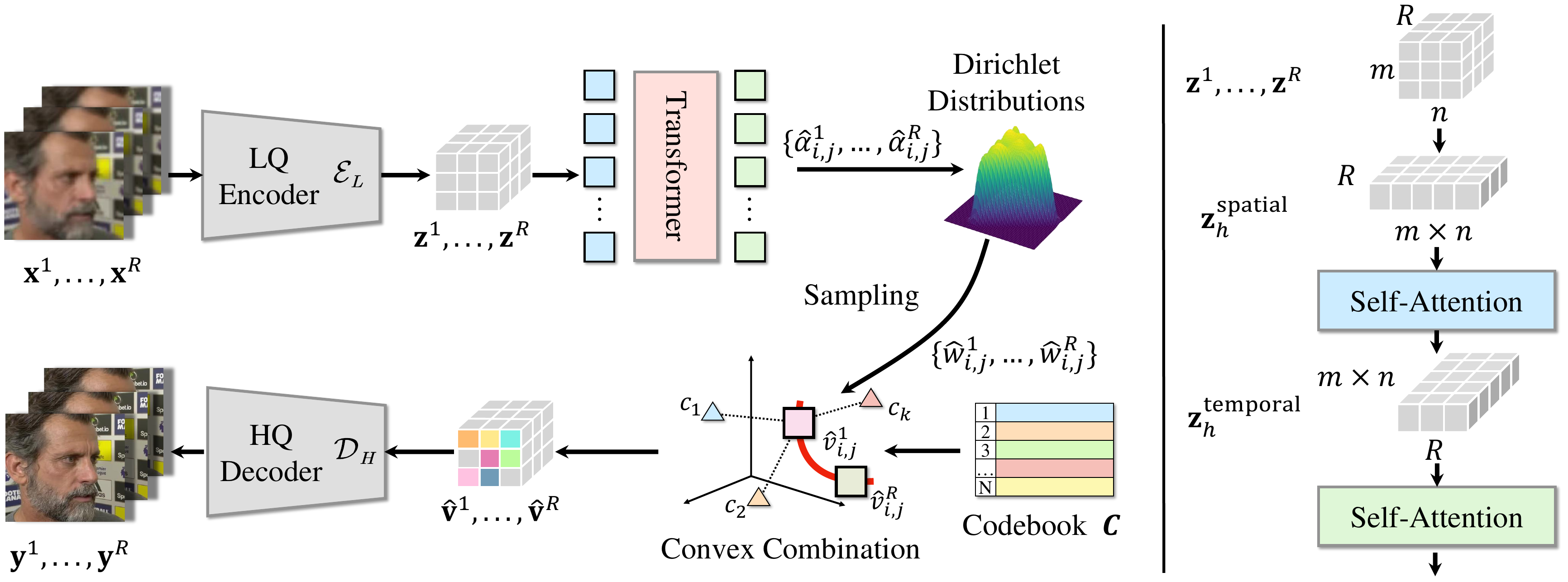}
    \vspace{-5mm}
    \caption{Overview of our framework. The framework processes a sequence of low-quality frames using three core components: 
(1) an encoder network that extracts spatial features; 
(2) a spatio-temporal Transformer that models inter-frame dependencies and predicts Dirichlet parameters for latent code distributions; 
and (3) a decoder that reconstructs high-quality frames from convex combinations of learnable codebook items. 
Crucially, the latent representation at each spatial location is formulated as a probabilistic mixture of codebook entries, enabling smooth transitions via variational inference over the Dirichlet manifold. 
This continuous relaxation is regularized by an ELBO objective, balancing reconstruction fidelity with temporal coherence.}
    \label{fig:pipeline}
    \vspace{-3mm}
\end{figure*}

\section{Related Works}
\label{sec:related}

\paragraph{Image Face Restoration.}
The objective of face image restoration is to recover high-quality facial images from low-quality inputs undergoing complex degradations.  
Existing methods primarily rely on three types of facial priors to mitigate dependence on degraded inputs.
Geometric priors~\cite{chen2018fsrnet, yu2018face} capture morphological constraints using landmarks~\cite{zhang2014facial, kazemi2014one, kowalski2017deep}, parsing maps~\cite{chen2021progressive, shen2018deep}, and component heatmaps~\cite{bulat2016human}.
Reference priors~\cite{li2018learning, li2020blind}, drawing on high-quality examples, facilitate fine-detail restoration and identity preservation (e.g., DFDNet~\cite{liu2015dfdnet} builds a facial component dictionary from VGGFace~\cite{parkhi2015deep}). 
Generative priors further enhance restoration by identifying optimal latent codes (PULSE~\cite{menon2020pulse}) or by incorporating pretrained StyleGAN~\cite{karras2020analyzing, karras2019style} into encoder-decoder frameworks (GPEN~\cite{yang2021gan}, GFP-GAN~\cite{wang2021towards}). 
Pretrained vector-quantization codebooks are also adopted to refine fine-grained details, as demonstrated by VQFR~\cite{gu2022vqfr} and CodeFormer~\cite{zhou2022towards}. 
More recently, DR2~\cite{wang2023dr2} and DifFace~\cite{yue2024difface} leverage diffusion models to remove degradations with strong fidelity and robustness.
However, directly extending image-based restoration methods to the video domain often fails to ensure temporal consistency of reconstructed facial details. 
Concurrently, there is a strong motivation to exploit pretrained prior distributions from single-image restoration—such as high-quality facial codebooks—in video scenarios. 
This work is driven by the objective of effectively harnessing these priors for consistent and robust video face restoration.
\vspace{-5mm}

\paragraph{General Video Restoration.}
Video restoration aims to enhance degraded video content to near-lossless quality. FSTRN~\cite{li2019fast} employs fast spatio-temporal networks with 3D convolutions for feature alignment and motion extraction. 
EDVR~\cite{wang2019edvr} leverages space-time deformable convolutions to aggregate temporal information across adjacent frames. 
RSDN~\cite{isobe2020video} splits inputs into structural and detail components, passing them through recurrent two-stream modules to refine texture details. 
BasicVSR~\cite{chan2021basicvsr} and BasicVSR++~\cite{chan2022basicvsr++} utilize optical flow for temporal alignment and integrate bidirectional hidden states from past and future frames, achieving state-of-the-art performance through efficient sequence-wide information fusion.
However, when these general video restoration methods are adapted (e.g., via data-driven fine-tuning or domain adaptation) to video face restoration, the lack of face-specific priors—such as facial codebooks, geometric constraints, and reference priors—often results in suboptimal detail recovery and diminished visual fidelity in facial regions.
\vspace{-5mm}
\paragraph{Codebook Based Learning.}
Vector-quantized (VQ) codebooks were first introduced in VQ-VAE~\cite{van2017neural}, where the codebook is learned during training rather than predefined. 
VQ-GAN~\cite{esser2021taming} subsequently incorporates perceptual and adversarial losses, enabling a more compact codebook while preserving or improving expressiveness and visual fidelity. 
Recent studies~\cite{yu2021vector,lancucki2020robust} further optimize codebook usage through L2 normalization and edge-prior regularization. 
Learned sparse representations have shown significant benefits in image restoration tasks (e.g., super-resolution~\cite{gu2015convolutional,timofte2015a+} and denoising~\cite{elad2006image}), offering higher efficiency and flexibility compared to traditional manually crafted dictionaries~\cite{jo2021practical,li2020blind}.
In the face restoration domain, a high-quality codebook pretrained on extensive high-resolution facial data provides a powerful face-specific prior, with quantization helping to reduce representational ambiguity. 
To address the inherent limitations of discrete quantization, we relax the latent space constraint, learning and predicting a continuous latent distribution for enhanced reconstruction.
\vspace{-3mm}
\section{Methodology}
\label{sec:method}

This section presents our methodology for video face restoration through five components: 
Section~\ref{subsec:prelim} establishes fundamental concepts in vector quantization and Dirichlet distributions. 
Section~\ref{subsec:continuous} details our continuous latent space formulation using variational Dirichlet inference. 
Section~\ref{subsec:loss} describes the hybrid loss function combining evidence lower bound optimization with perceptual metrics. 
Section~\ref{subsec:architecture} outlines the spatial-temporal transformer architecture, while Section~\ref{subsec:training} specifies training protocols and inference strategies for temporal coherence.
The framework overview is shown in Figure~\ref{fig:pipeline}.
\vspace{-1mm}
\subsection{Preliminary}\label{subsec:prelim}
\paragraph{Vector-Quantized Autoencoder.}
The vector-quantized autoencoder framework aims to reconstruct a high-quality image $\mathbf{y} \in \mathbb{R}^{H \times W \times 3}$ from its degraded counterpart $\mathbf{x} \in \mathbb{R}^{H \times W \times 3}$. 
The architecture comprises four components: an encoder $\mathcal{E}_L$, a decoder $\mathcal{D}_H$, a Transformer network $\mathcal{H}$, and a learnable codebook $\mathbf{c} = [c_k]_{k=1}^N$ with $c_k \in \mathbb{R}^d$. 

Specifically, the encoder $\mathcal{E}_L$ first maps $\mathbf{x}$ to a latent feature map $\mathbf{z} \in \mathbb{R}^{m \times n \times d}$. 
The Transformer $\mathcal{H}$ then processes $\mathbf{z}$ to infer a probability distribution $\mathbf{s} \in \mathbb{R}^{m \times n \times N}$ over the codebook, where each spatial location $(i,j)$ satisfies:
\begin{equation}
\sum_{k=1}^{N} s_{i,j,k} = 1 \quad\text{and}\quad s_{i,j,k} \geq 0.
\end{equation}
The quantized feature map $\mathbf{\hat{z}}$ is obtained by replacing each $\mathbf{z}_{i,j}$ with the codebook entry $c_k$ corresponding to $\arg\max_k s_{i,j,k}$. 
Finally, the decoder $\mathcal{D}_H$ reconstructs $\mathbf{y}$ from $\mathbf{\hat{z}}$. 

This discrete quantization reduces representational ambiguity by constraining the latent space to the codebook entries. However, in video processing, hard quantization introduces temporal inconsistencies that manifest as flicker artifacts across reconstructed frames.

\paragraph{Dirichlet Distribution.}
A Dirichlet distribution is a continuous multivariate distribution defined on the $(N-1)$-dimensional simplex. 
For a parameter vector $\boldsymbol{\alpha} = [\alpha_1, \ldots, \alpha_N]$ with $\alpha_k > 0$, its probability density function for $\mathbf{w} = [w_1, \ldots, w_N]$ is given by
\begin{align}
\mathrm{Dir}(\mathbf{w} \mid \boldsymbol{\alpha}) = \frac{\Gamma\bigl(\sum_{k=1}^{N}\alpha_k\bigr)}{\prod_{k=1}^{N}\Gamma(\alpha_k)} 
\prod_{k=1}^{N} w_k^{\alpha_k - 1},\\
\text{subject to } \sum_{k=1}^{N} w_k = 1 \text{ and } w_k \ge 0.
\end{align}
Here, $\Gamma(\cdot)$ denotes the Gamma function. 
As the conjugate prior for categorical distributions, the Dirichlet provides a probabilistic framework for modeling uncertainty in discrete classifications. 
In our approach, it enables continuous relaxation of latent code assignments by representing each spatial location's latent code as a convex combination of codebook entries. 
This probabilistic formulation permits smooth transitions between frames through variational inference over the Dirichlet manifold, effectively mitigating temporal artifacts such as flicker in video reconstructions.

\subsection{Continuous Latent Space}\label{subsec:continuous}
To overcome the limitations of discrete quantization, we relax the latent space constraint by allowing it to span the convex hull of the codebook $\{c_k\}_{k=1}^N$.
Specifically, each latent code $\mathbf{\hat{v}} = [\hat{v}_{i,j}]_{m\times n}$ is modeled as a convex combination of the code items, weighted by $\mathbf{\hat{w}} = [\hat{w}_{i,j}]_{m\times n}$:
\begin{equation}
\hat{v}_{i,j} = \hat{w}_{i,j}^T\mathbf{c},
\label{eq:latent_code}
\end{equation}
where $\hat{w}_{i,j} \in \mathbb{R}^N$, $\hat{w}_{i,j} \ge \mathbf{0}$, and $\hat{w}_{i,j}^T\mathbf{1} = 1$. 
This formulation ensures that each $\hat{v}_{i,j}$ resides within the convex hull of the learned code items, thereby providing a continuum of feasible latent representations.

We adopt a variational approach to learn and predict the distribution of $\mathbf{\hat{w}}$ via neural networks. 
Specifically, we approximate the true posterior distribution $p(\mathbf{\hat{w}} \mid \mathbf{x})$ by $q_{\theta}(\mathbf{\hat{w}} \mid \mathbf{x})$, where $\theta$ denotes the network parameters. 
We further assume:
\begin{equation}
    q_{\theta}(\mathbf{\hat{w}} \mid \mathbf{x}) = \mathtt{Dir}(\mathbf{\hat{\alpha}} \mid \mathbf{x}),
\end{equation}
where $\mathtt{Dir}$ denotes the Dirichlet distribution and $\mathbf{\hat{\alpha}} = [\hat{\alpha}_{i,j}]_{m\times n}$ are the Dirichlet hyper-parameter satisfying $\hat{\alpha}_{i,j} \in \mathbb{R}^N$ and $\hat{\alpha}_{i,j} \ge 0$. 

To learn $\theta$, we maximize the evidence lower bound (ELBO):
\begin{equation}
    \log p_{\theta}(\mathbf{y} \mid \mathbf{x}) \ge \mathcal{L}_{\mathtt{ELBO}}(\theta; \mathbf{x}, \mathbf{y}),
\end{equation}
where
\begin{align}
    \mathcal{L}_{\mathtt{ELBO}}(\theta;\mathbf{x},\mathbf{y})
    = &-\mathtt{KL}\Bigl(q_{\theta}(\mathbf{\hat{w}} \mid \mathbf{x}) \Big\| p_{\theta}(\mathbf{\hat{w}})\Bigr) \nonumber \\
    &+ \mathbb{E}_{q_{\theta}(\mathbf{\hat{w}} \mid \mathbf{x})}\bigl[\log p_{\theta}(\mathbf{y}\mid \mathbf{x}, \mathbf{\hat{w}})\bigr].
\label{eq:elbo}
\end{align}
The first term is the Kullback–Leibler (KL) divergence between the approximate posterior $q_{\theta}(\mathbf{\hat{w}} \mid \mathbf{x})$ and the prior $p_{\theta}(\mathbf{\hat{w}})$. Assuming $p_{\theta}(\mathbf{\hat{w}}) = \mathtt{Dir}(\alpha)$, each KL term can be derived in closed form:
\begin{align}
    \mathtt{KL}(\cdot \| \cdot)
    &= \mathtt{C}
    + \log \Gamma\Bigl(\sum_k \hat{\alpha}_{i,j,k}\Bigr)
    -\sum_k \log \Gamma\bigl(\hat{\alpha}_{i,j,k}\bigr) 
    \nonumber \\
    &\quad + \sum_k \Bigl(\hat{\alpha}_{i,j,k}-\alpha_{i,j,k}\Bigr) \Bigl[\psi\bigl(\hat{\alpha}_{i,j,k}\bigr) - \psi\Bigl(\sum_k \alpha_{i,j,k}\Bigr)\Bigr],
\label{eq:kl}
\end{align}
where $\mathtt{C}$ is a constant and $\psi(\cdot)$ denotes the digamma function. When all the items in $\hat{\alpha}_{i,j}$ are close to zero, the sampled weight vector $\hat{w}_{i,j}$ tends to be a one-hot vector. 
In such cases, only the nearest code items in the codebook would have impact on the latent code $\hat{v}_{i,j}$. Conversely, if all the items in $\hat{\alpha}_{i,j}$ are larger, the sampled weight vector tends to be uniform, and the latent code $\hat{v}_{i,j}$ would be an uniform average of the code items in the codebook. 

The second term is the expected reconstruction error, and would encourage the sampled weight vector $\hat{w}_{i,j}$ to produce a latent code that can recover the high-quality image $\mathbf{y}$ as close as possible by the decoder network $\mathcal{D}_{H}$. 

\subsection{Loss Function}\label{subsec:loss}
Given the prediction $\mathbf{y}^{\mathtt{pred}}$ and the ground truth $\mathbf{y}$, the overall loss function combines the evidence lower bound (ELBO) and the learned perceptual image patch similarity (LPIPS) loss as follows:
\begin{equation}
    \mathcal{L}_{\text{total}} = \lambda_1 \mathcal{L}_{\mathtt{ELBO}} + \lambda_2 \mathcal{L}_{\mathtt{LPIPS}},
\end{equation}
where $\lambda_1$ and $\lambda_2$ balance the two terms.
During training, we set $\lambda_1=-1.0$ and $\lambda_2=1.0$ to prioritize reconstruction fidelity while maintaining perceptual quality.

\paragraph{ELBO Loss.} 
We compute the ELBO following Eq.~(\ref{eq:elbo}). 
Specifically, the KL divergence term is computed following Eq.~(\ref{eq:kl}) to encourage the predicted $\hat{\alpha}$ to be close to the prior $\alpha$. 
For the expected reconstruction error term, we approximate the expectation by the Monte Carlo method. Moreover, we assume the $\log p_{\theta}(\mathbf{y}\mid \mathbf{x}, \mathbf{\hat{w}})$ follows the Laplacian distribution whose location parameters are $\mathbf{y}^{\mathtt{pred}}$ and the scale parameters are 1. Accordingly, the expected reconstruction error term is computed as follows:
\begin{align}
\mathbb{E}_{q_{\theta}(\mathbf{\hat{w}} \mid \mathbf{x})}&\bigl[\log p_{\theta}(\mathbf{y}\mid \mathbf{x}, \mathbf{\hat{w}})\bigr]  \approx \sum_{\mathbf{\hat{w}}_l\sim q_{\theta}(\mathbf{\hat{w}} \mid \mathbf{x})} \log p_{\theta}(\mathbf{y}\mid \mathbf{x}, \mathbf{\hat{w}}_l) \nonumber\\
& = \sum_{\mathbf{\hat{w}}_l\sim q_{\theta}(\mathbf{\hat{w}} \mid \mathbf{x})} |\mathbf{y}-\mathbf{y}^{\mathtt{pred}}_l| + \mathtt{C}
\end{align}
where $\mathbf{\hat{w}}_l\sim q_{\theta}(\mathbf{\hat{w}} \mid \mathbf{x})$ means differentiably sampling $\mathbf{\hat{w}}_l$ from the distribution $q_{\theta}(\mathbf{\hat{w}} \mid \mathbf{x})$, $\mathbf{y}^{\mathtt{pred}}_l$ is the predicted high-quality image when the sampled weight is $\mathbf{\hat{w}}_l$, and $\mathtt{C}$ is a constant.

\paragraph{Perceptual Loss.} We incorporate LPIPS~\cite{zhang2018perceptual} to enhance visual quality, computed as:
\begin{equation}
    \mathcal{L}_{\mathtt{LPIPS}} = \|\phi(\mathbf{y}) - \phi(\mathbf{y}^{\mathtt{pred}})\|_2^2,
\end{equation}
where $\phi$ denotes deep features from a pretrained VGG network.

\begin{figure*}[!t]
    \centering
    \includegraphics[width=0.97\textwidth]{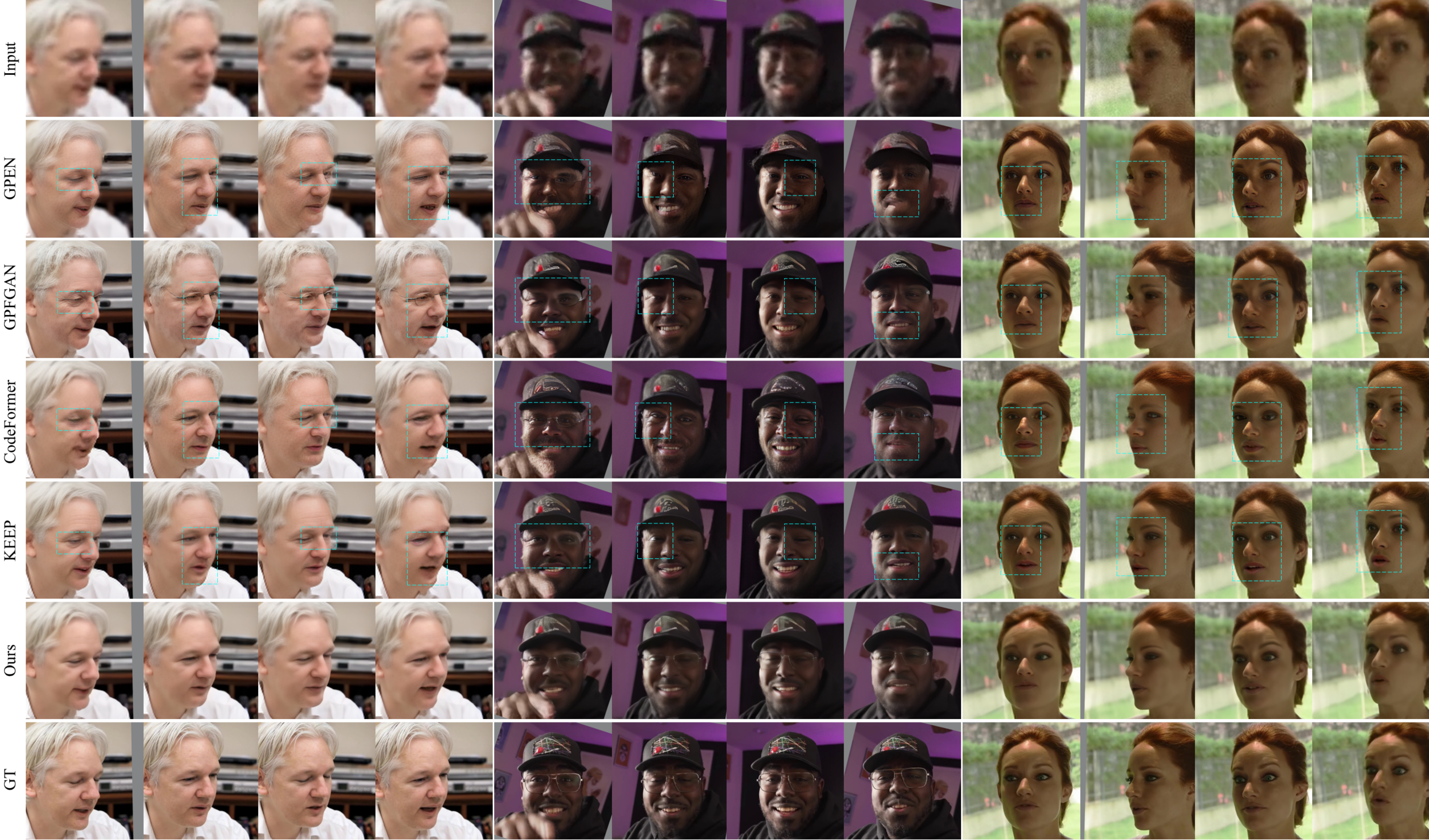}
    \vspace{-3mm}
    \caption{Qualitative comparison with other state-of-the-art methods on the VFHQ-Test dataset for blind face restoration. Our model demonstrates superior performance in recovering finer details and achieving significantly better temporal consistency, particularly under challenging conditions such as large facial angles, body occlusions, and substantial facial movements.}
    \vspace{-1mm}
    \label{fig:QualitativeComparison}
\end{figure*}

\begin{table*}[htbp]
    \centering
    \begin{tabular}{@{}c|cccccccc@{}}
        \toprule
        Method     & PSNR↑                & SSIM↑       & \multicolumn{1}{r}{LPIPS↓} & \multicolumn{1}{r}{IDS↑} & \multicolumn{1}{r}{AKD↓}   & \multicolumn{1}{r}{FVD↓}    & TLME↓  \\ \midrule
        GPEN           & 26.509     & 0.739     & 0.341       & 0.856     & 2.920     & 405.926       & 1.641         \\
        GFPGAN         & 27.221     & 0.775     & 0.311       & 0.861     & 2.998     & 359.197       & 1.223         \\
        CodeFormer     & 26.064     & 0.740     & 0.320       & 0.781     & 3.479     & 510.034       & 1.530         \\ \midrule
        RealBasicVSR   & 26.030     & 0.715     & 0.407       & 0.811     & 3.181     & 635.216       & 1.777         \\
        BasicVSR++     & 27.001     & 0.775     & 0.409       & 0.826     & 3.513     & 823.908       & 1.598         \\ \midrule
        PGTFormer      & 27.829     & 0.786     & 0.292       & 0.879     & 2.566     & \textbf{332.340} & 1.333  \\
        KEEP           & 27.810     & 0.797     & 0.268       & 0.863     & 2.466     & 378.72        & 1.156        \\ \midrule
        Ours  & \textbf{29.099}     &   \textbf{0.831}      &    \textbf{0.246}  &   \textbf{0.908}     &     \textbf{2.093}   &    336.015    &           \textbf{1.091}    \\
        \bottomrule
    \end{tabular}
    \vspace{-3mm}
    \caption{Quantitative comparison on the VFHQ-Test dataset for blind video face restoration.}
    \vspace{-4mm}
    \label{tab:comparison_sota_vfr}
\end{table*}

\subsection{Network Architecture}\label{subsec:architecture}
The proposed architecture processes a sequence of $R$ low-quality frames $\{\mathbf{x}^r\}_{r=1}^R$ to restore their high-quality counterparts $\{\mathbf{y}^r\}_{r=1}^R$, as illustrated in Figure~\ref{fig:pipeline}. 
The framework comprises three key components:
\paragraph{Encoder.} 
Each input frame $\mathbf{x}^r$ is first encoded into a latent feature map $\mathbf{z}^r \in \mathbb{R}^{m \times n \times d}$ via a convolutional encoder $\mathcal{E}_L$. 
The encoder reduces spatial resolution through five strided convolutions, yielding $m = H/32$ and $n = W/32$ for input resolution $H \times W$.
\paragraph{Spatio-Temporal Transformer.} 
The Transformer $\mathcal{H}$ with $2H$ alternating attention blocks then process the sequence $\{\mathbf{z}^1,\ldots,\mathbf{z}^R\}$.  
For odd-indexed blocks $h \in \{1,3,...,2H-1\}$, spatial self-attention is applied by reshaping features into $\mathbf{Z}_h^\mathtt{spatial} \in \mathbb{R}^{R \times (mn) \times d}$ and computing:  
\begin{equation}
\mathbf{Z}_{h+1}^\mathtt{spatial} = \mathtt{Softmax}\left(\frac{Q_h K_h^T}{\sqrt{d}}\right)V_h + \mathbf{Z}_h^\mathtt{spatial},
\end{equation}
where $Q_h, K_h, V_h \in \mathbb{R}^{(mn) \times d}$ are query, key, and value projections. For even-indexed blocks, temporal attention operates on $\mathbf{Z}_h^\mathtt{temporal} \in \mathbb{R}^{(mn) \times R \times d}$ via analogous computations across the temporal dimension. 
Sinusoidal positional embeddings augment queries and keys to encode spatial/temporal positions.

\paragraph{Code Prediction \& Decoder.} 
The Transformer output $\{\mathbf{z}^r\}_{r=1}^R$ is linearly projected to predict Dirichlet parameters $\{\hat{\boldsymbol{\alpha}}^r\}_{r=1}^R$. For each frame, latent codes $\hat{\mathbf{v}}^r$ are sampled by $\hat{\mathbf{w}}^r \sim \mathtt{Dir}(\hat{\boldsymbol{\alpha}}^r)$ and computed as $\hat{\mathbf{v}}^r = \sum_{k=1}^N \hat{w}_{i,j,k}^r c_k$ per spatial location. 
These codes are decoded to $\mathbf{y}^r$ through a mirrored decoder $\mathcal{D}_H$ with five transposed convolutions.
\vspace{-5mm}
\paragraph{Implementation Details.} 
The encoder/decoder each contain 12 residual blocks and 5 resolution-scaling layers. 
The Transformer employs $H=4$ alternating spatial-temporal blocks (8 total) with 8 attention heads. 
We evaluate codebook sizes $N \in \{256, 512, 1024\}$, with $d=256$ codes. 
During training, we progressively unfreeze components: first the encoder/decoder, then the Transformer, and finally the codebook.
\vspace{-1mm}
\subsection{Training and Inference}\label{subsec:training}
\paragraph{Training.} 

The model parameters were initialized using a combination of pre-trained weights and random initialization. To investigate the impact of codebook size on model performance, we retrained CodeFormer~\cite{liu2023codeformer} with codebook sizes of 256 and 512 on the FFHQ dataset~\cite{karras2019style}, which consists of aligned and resized facial images at a resolution of 512x512. During the model parameter initialization phase, we loaded the pre-trained weights of CodeFormer. To ensure temporal stability in the task of video face restoration, we incorporated a 9-layer Multi-Head Attention (MHA) Temporal Transformer, whose parameters were randomly initialized based on a Gaussian distribution. Throughout the training process, the codebook parameters remained frozen. The newly added Temporal Transformer parameters were consistently trained, and we also experimented with freezing the encoder and decoder to assess their impact on performance metrics and results. The experiments demonstrated that allowing the encoder and decoder parameters to participate in training yielded superior image quality and temporal stability.
\vspace{-1mm}
\paragraph{Inference.} 
During inference, we process input videos using a sliding window approach with a stride of 1 frame. 
Each window consists of 5 consecutive frames, padded at the sequence boundaries by replicating the initial and final frames. 
The network predicts the restored central frame (third position) for each window. Final video reconstructions are obtained by aggregating these center frame predictions, ensuring temporal coherence through overlapping window processing. 
This strategy effectively balances computational efficiency with temporal consistency preservation.

\begin{table*}[ht]
    \centering
    \begin{tabular}{l *{7}{c@{\hspace{1pt}}c}} 
        \toprule
        Methods & 
        \multicolumn{2}{c}{PSNR$\uparrow$} & 
        \multicolumn{2}{c}{SSIM$\uparrow$} & 
        \multicolumn{2}{c}{LPIPS$\downarrow$} & 
        \multicolumn{2}{c}{IDS$\uparrow$} & 
        \multicolumn{2}{c}{AKD$\downarrow$} & 
        \multicolumn{2}{c}{FVD$\downarrow$} & 
        \multicolumn{2}{c}{TLME$\downarrow$} \\
        \cmidrule(lr){2-3} \cmidrule(lr){4-5} \cmidrule(lr){6-7} \cmidrule(lr){8-9} \cmidrule(lr){10-11} \cmidrule(lr){12-13} \cmidrule(lr){14-15}
        & {inp.} & {col.} & {inp.} & {col.} & {inp.} & {col.} & {inp.} & {col.} & {inp.} & {col.} & {inp.} & {col.} & {inp.} & {col.} \\
        \midrule
        GPEN    & 28.170/ & 24.364 & 0.919/ & 0.945 & 0.156/ & 0.204 & 0.902/ & 0.999 & 2.742/ & 1.350 & 649.8/ & 212.1 & 4.779/ & 1.181 \\
        CodeFormer & 31.463/ & 18.505 & 0.949/ & 0.682 & 0.144/ & 0.457 & 0.927/ & 0.711 & 2.645/ & 9.082 & 224.3/ & 826.5 & 1.744/ & 3.264 \\
        PGDiff  & 23.146/ & 21.735 & 0.853/ & 0.862 & 0.236/ & 0.388 & 0.859/ & 0.973 & 5.929/ & 2.111 & 590.7/ & 470.2 & 1.831/ & 1.667 \\
        \midrule
        Ours    & \textbf{31.630}/ & \textbf{26.885} & \textbf{0.953}/ & \textbf{0.962} & \textbf{0.069}/ & \textbf{0.141} & \textbf{0.945}/ & \textbf{0.999} & \textbf{1.379}/ & \textbf{0.907} & \textbf{147.7}/ & \textbf{155.7} & \textbf{1.290}/ & \textbf{0.964} \\
        \bottomrule
    \end{tabular}
    \vspace{-3mm}
    \caption{Quantitative comparison on the VFHQ-Test for video face inpainting and colorization. ``inp.'': inpainting, ``col.'': colorization.}
    \label{tab:inpaint_color}
    \vspace{-3mm}
\end{table*}

\section{Experiments}\label{sec:experiments}
\subsection{Experimental Settings}
\paragraph{Datasets.}
The proposed method is trained utilizing the VFHQ~\cite{xie2022vfhq} dataset, comprising 16000 video clips with native 512$\times$512 resolution. 

For blind face restoration (BFR) task, we adopt the degradation pipeline from VFHQ~\cite{xie2022vfhq}. We put the details in Appendix. For the colorization task, we converted the videos to grayscale, and for the inpainting task, we applied brush stroke masks~\cite{yang2021gan} to randomly draw irregular polyline masks for generating masked faces. During evaluation, facial alignment preprocessing is applied to test samples to ensure compatibility with baseline methods requiring geometrically normalized inputs.

\paragraph{Settings and Metrics.} We introduce the settings and metrics in Appendix.

\subsection{Comparisons with State-of-the-Art Methods}
\paragraph{Restoration}
For the video face restoration task, we perform comparisons with state-of-the-art methods, including KEEP~\cite{feng2024kalman} and PGTFormer~\cite{xu2024beyond}. We also present the results of BasicVSR++~\cite{chan2022basicvsr++} and RealBasicVSR~\cite{chan2022investigating}, which are designed for general-purpose video restoration. The results of single-image face restoration frameworks including CodeFormer~\cite{liu2023codeformer}, GFPGAN~\cite{wang2021towards} and GPEN~\cite{yang2021gan} are also presented for comparison. The quantitative results are in Table~\ref{tab:comparison_sota_vfr}. We achieve the best performance under all the metrics except for the FVD. Especially, our framework reduced the TLME from 1.156 to 1.091, showing better temporal consistency. The visualization of the predictions is shown in Figure~\ref{fig:QualitativeComparison}. Our framework produces higher-quality images with richer details. 

\begin{figure}[]
    \centering
    \includegraphics[width=0.46\textwidth]{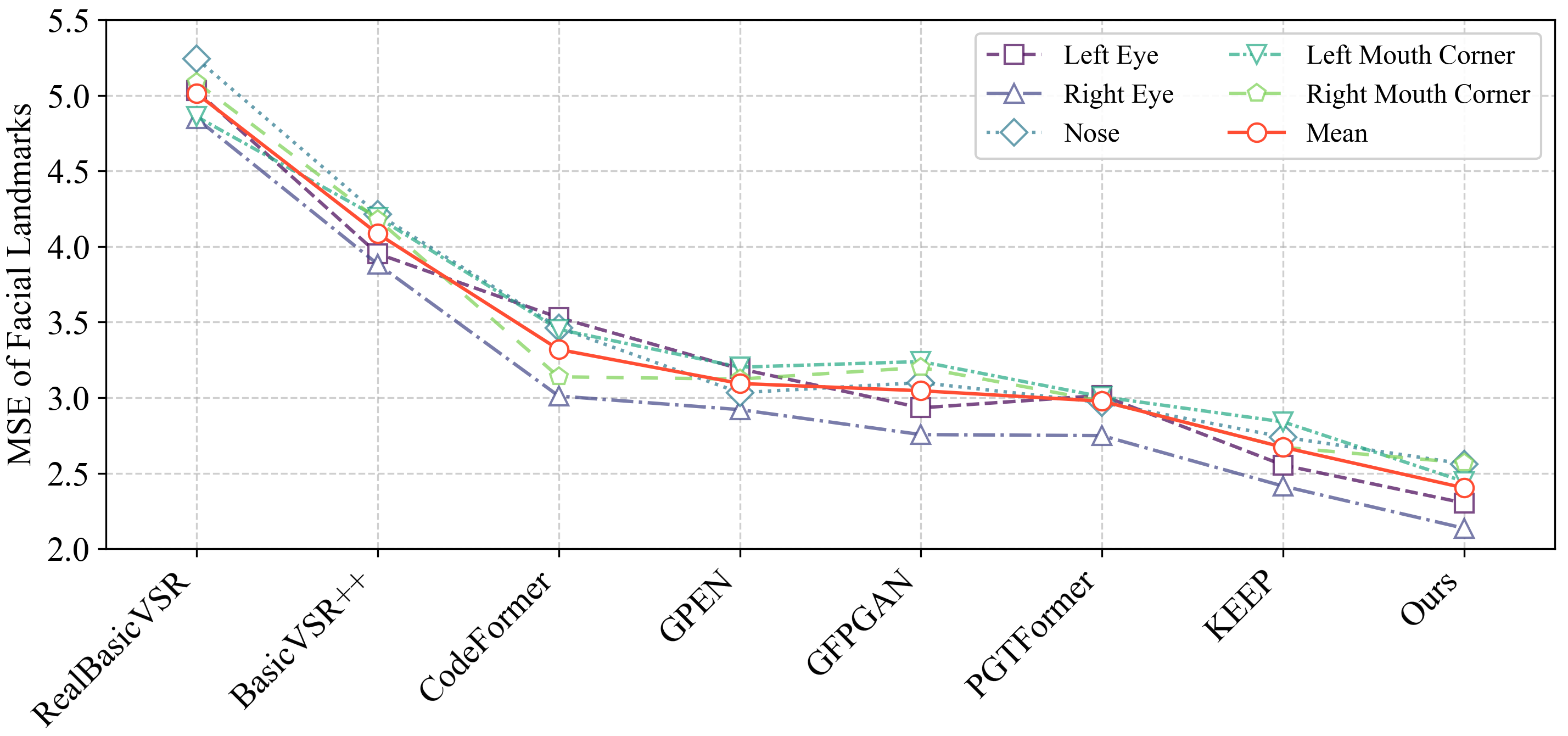}
    \vspace{-3mm}
    \caption{Comparison of temporal stability with other state-of-the-art methods on the VFHQ-Test dataset. In the context of blind face restoration, we calculate the discrepancies between the five facial landmarks of the output results and the ground truth for each method. These results indicate that our method achieves the lowest mean squared error (MSE) at each landmark, demonstrating the best temporal stability among all evaluated approaches.}
    \vspace{-4mm}
    \label{fig:facial_mse}
\end{figure}

\begin{figure}
    \centering
    \begin{subfigure}{0.46\textwidth}
    \includegraphics[width=1.0\linewidth]{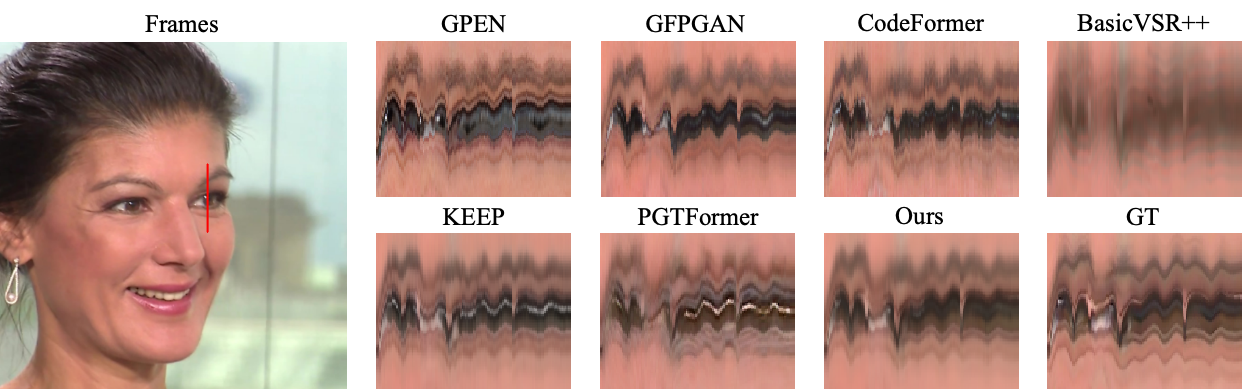}
    \end{subfigure}
    \vspace{-3mm}
    \caption{Comparison of temporal stability with other state-of-the-art methods for face restoration on the VFHQ-Test dataset. We selected a column (red line) along the subject's eye and plotted its temporal variations over time. Our method exhibits significantly mitigated temporal jitter, enhancing spatial consistency across restored frames and preserving temporal continuity over time.}
    \label{fig:temporal_stability}
\end{figure}

We also evaluate the temporal consistency in Figure~\ref{fig:facial_mse} and ~\ref{fig:temporal_stability}. In Figure~\ref{fig:facial_mse}, our predictions show lower errors at facial landmarks. In Figure~\ref{fig:temporal_stability}, our framework ensures temporal consistency and mitigates jitters across frames.

In addition, we evaluate the methods on out-of-domain data. The results and analysis are in Appendix.

\begin{figure}[]
    \centering
    \includegraphics[width=0.46\textwidth]{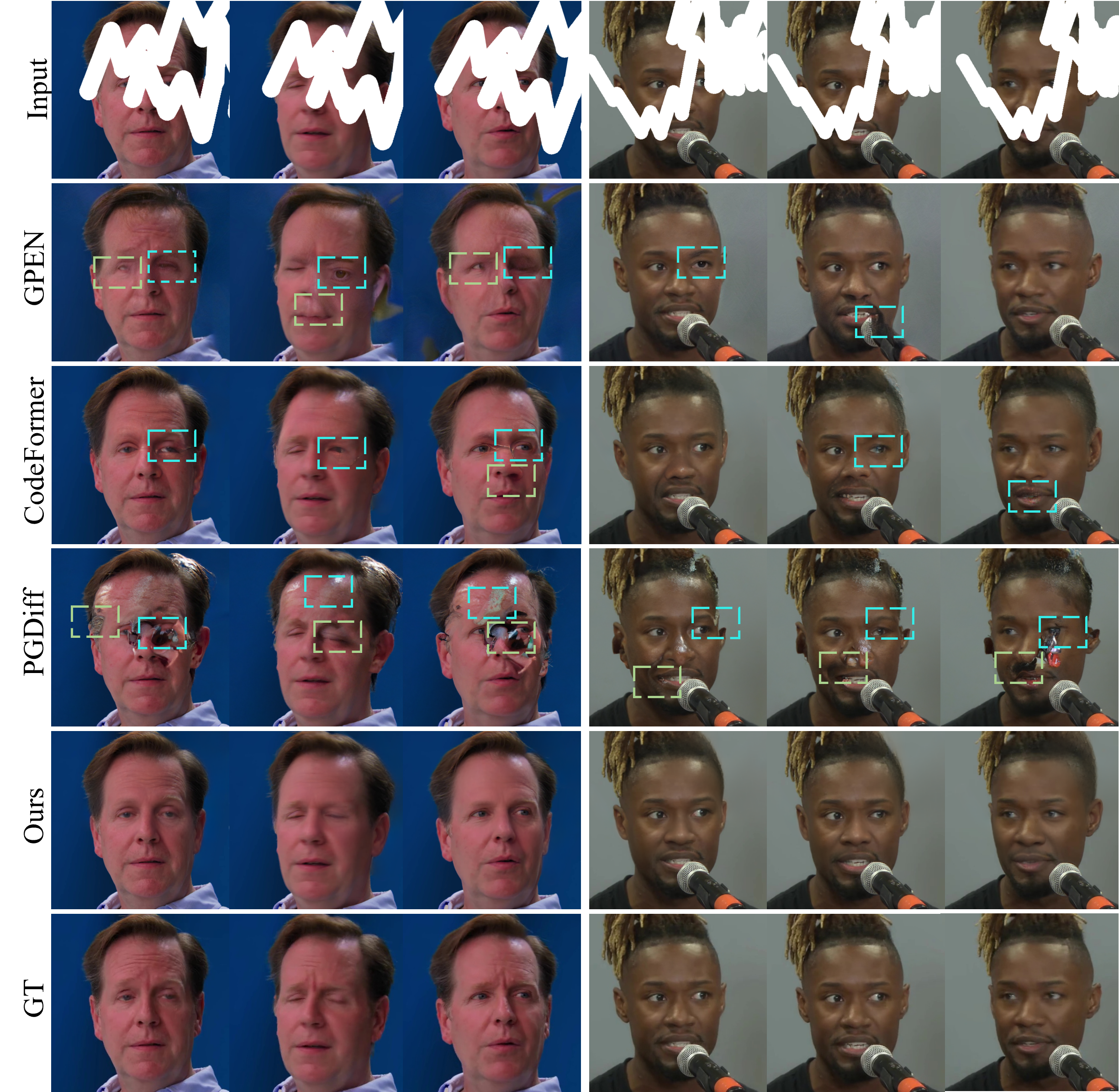}
    \vspace{-3mm}
    \caption{ Visual comparison with advanced inpainting methods on the VFHQ-Test dataset for challenging inpainting cases. CodeFormer and other methods struggle to restore details like eyes and lips. In contrast, our method is able to recover these details reasonably well, achieving a more realistic result.}
    \vspace{-5mm}
    \label{fig:inpainting_comparison}
\end{figure}

\begin{figure}[]
    \centering
    \includegraphics[width=0.46\textwidth]{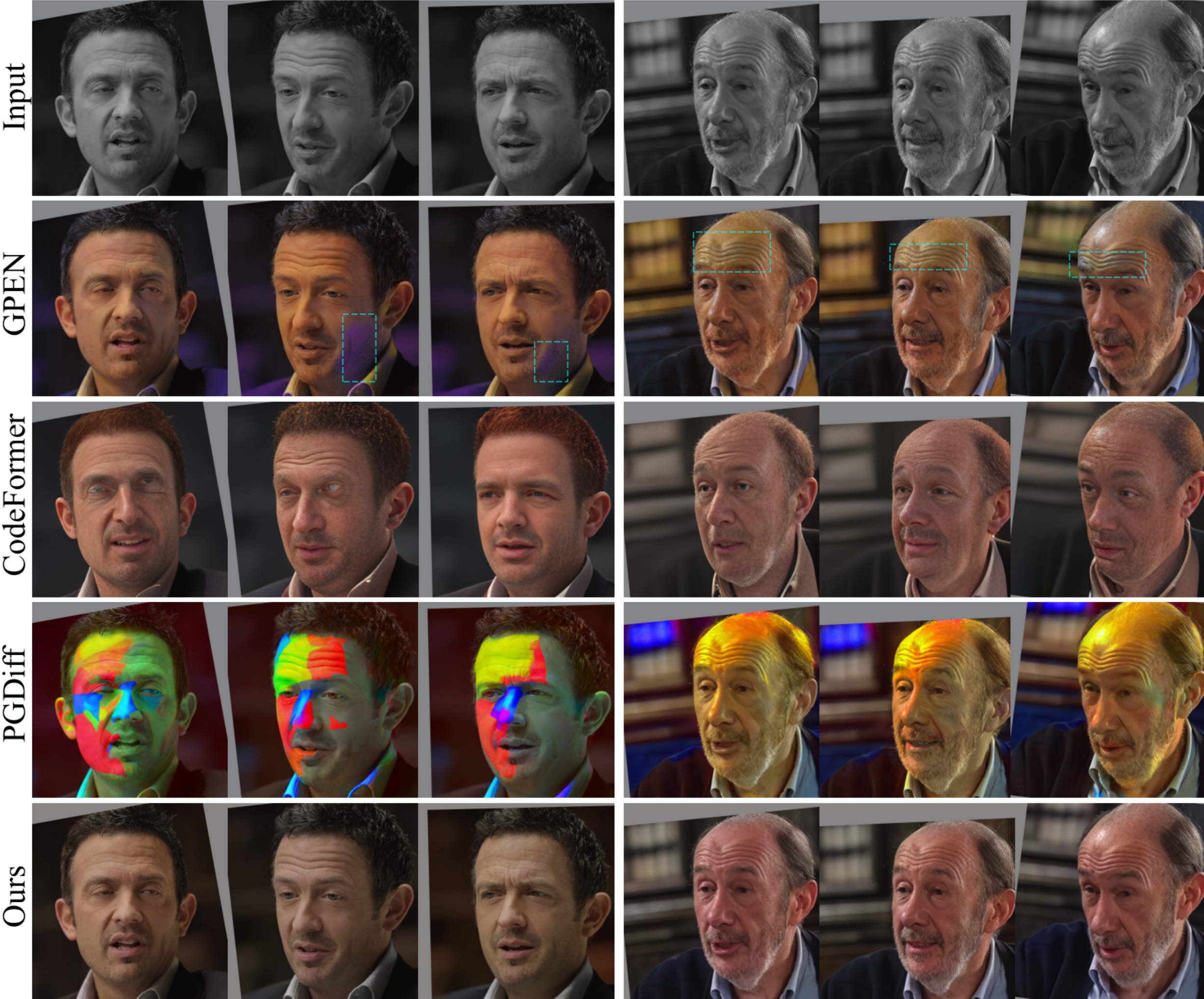}
    \vspace{-3mm}
    \caption{Visual comparison of colorization results on the VFHQ-Test dataset. Other methods yield unrealistic colorization, which include issues such as unnatural skin tones. In contrast, our method generates more realistic results, accurately capturing the natural hues and achieving a more lifelike appearance.}
    \vspace{-3mm}
    \label{fig:colorization_comparison}
\end{figure}



\paragraph{Inpainting}
In Table~\ref{tab:inpaint_color}, we compare with state-of-the-art frameworks for face inpainting, including GPEN~\cite{yang2021gan}, CodeFormer~\cite{liu2023codeformer}, and PGDiff~\cite{yang2023pgdiff}. We found our framework achieves much lower LPIPS, AKD and FVD, meaning that our prediction is closer to the ground truth under these metrics. In Figure~\ref{fig:inpainting_comparison} we present the predictions of different methods. Our framework produces correct eyes and mouses. We put more comparisons about the temporal variations in Appendix. 

\paragraph{Colorization}
We evaluate the colorization ability of our framework and compare it with GPEN~\cite{yang2021gan}, CodeFormer~\cite{liu2023codeformer}, and PGDiff~\cite{yang2023pgdiff}. The results in Table~\ref{tab:inpaint_color} show our framework achieves best performance under all the metrics. Especially, we reduce FVD from 212.075 to 155.727, and reduce TLME from 1.181 to 0.964, meaning that our predictions are closer to the ground truth and have better temporal consistency. In Figure~\ref{fig:colorization_comparison}, our framework produces more realistic colors. 
\subsection{Ablation Studies}
We perform ablations to demonstrate the effectiveness of our design. All the experiments are performed on VFHQ-Test dataset for the task of blind video face restoration. More experiments about the temporal Transformer, codebook size, and number of samples are in Appendix.

\begin{figure}
    \centering
    \begin{subfigure}{0.23\textwidth}
    \includegraphics[width=1.0\linewidth]{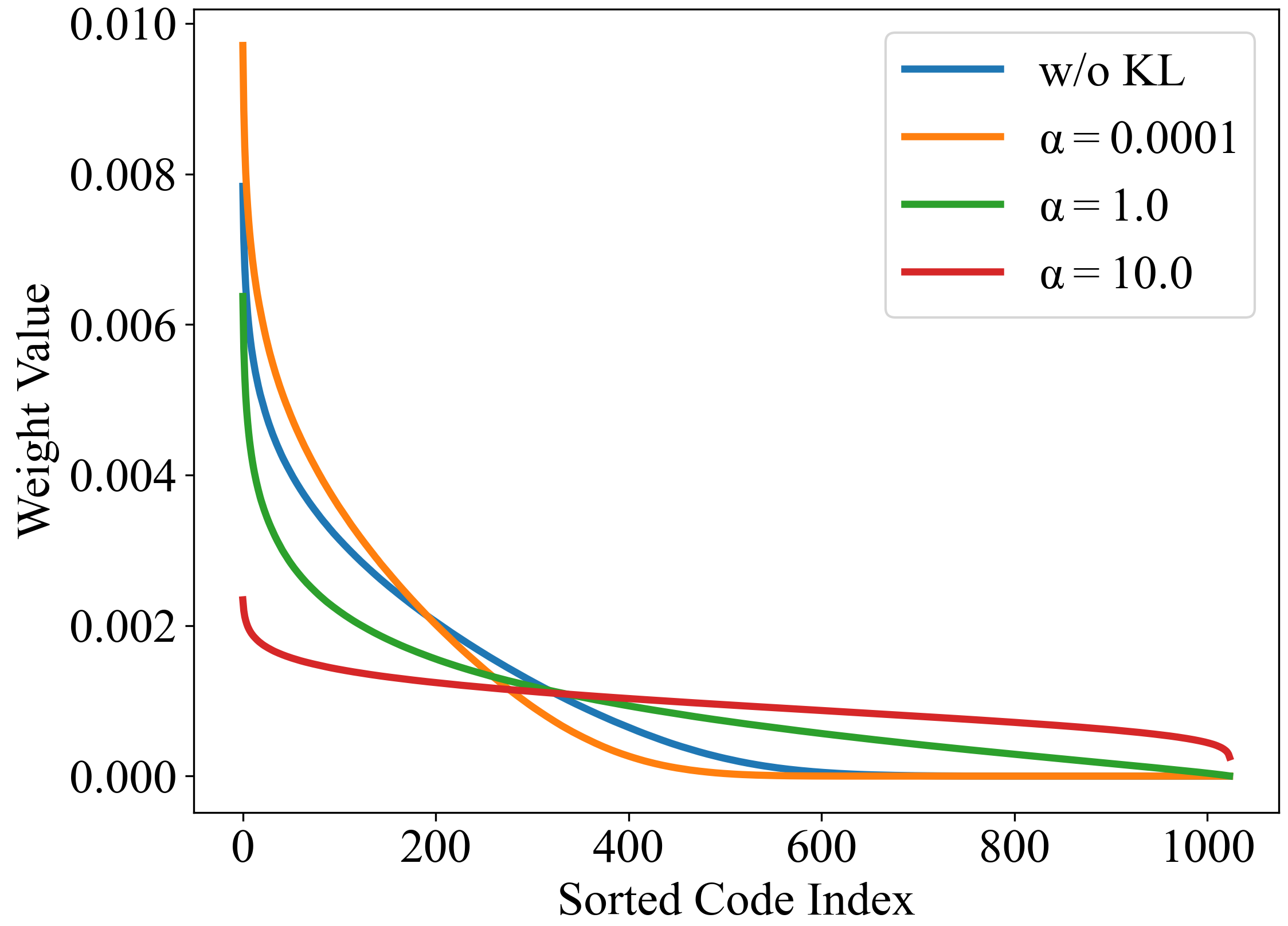}
    \caption{Weight values from different $\alpha$.}
    \end{subfigure}
    \begin{subfigure}{0.23\textwidth}
    \includegraphics[width=1.0\linewidth]{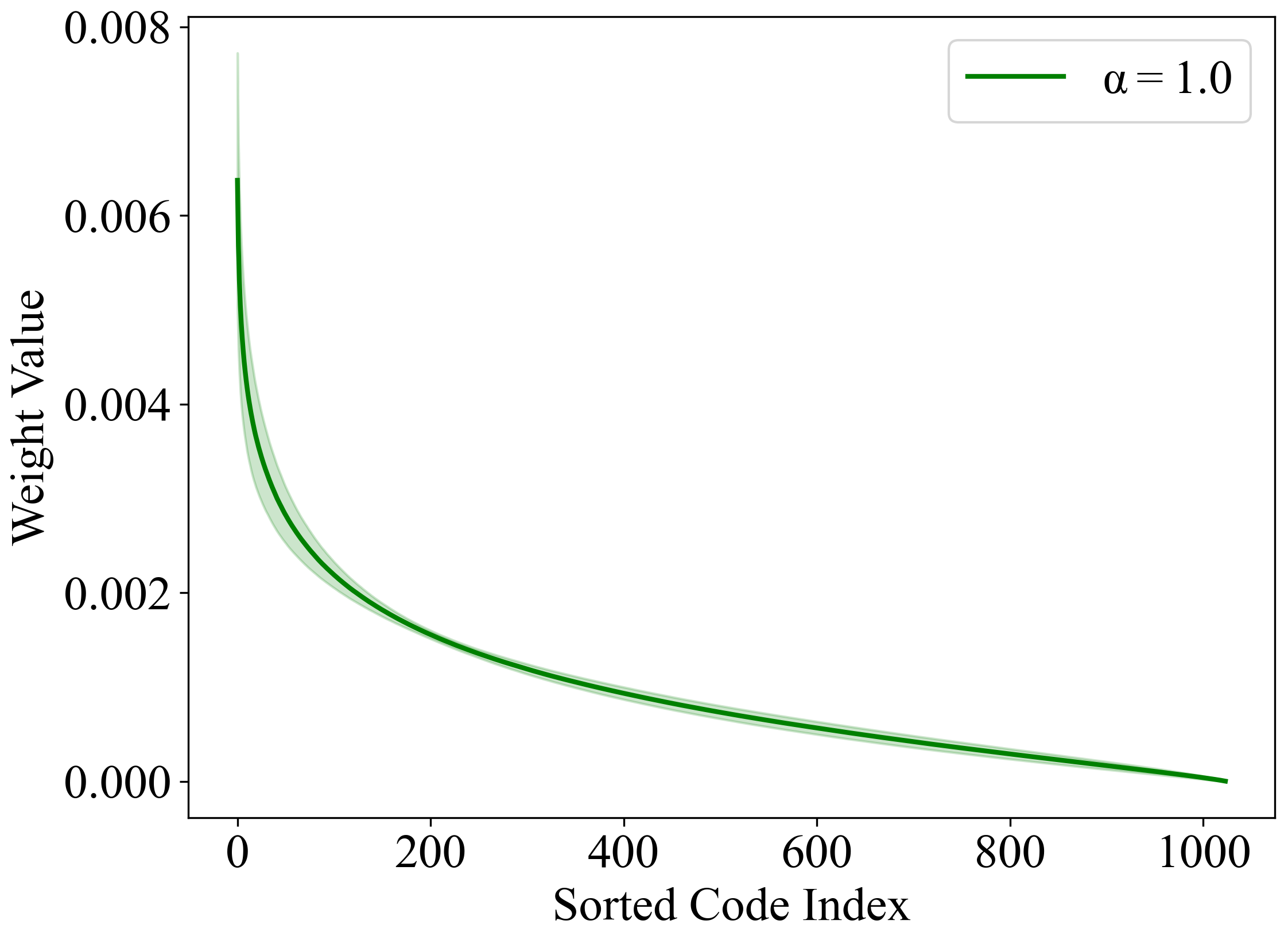}
    \caption{Variance of weight items.}
    \end{subfigure}
    \vspace{-3mm}
    \caption{Visualization of weight vector $\hat{w}_{i,j}$ which is used for averaging the code items. The weight items have been sorted in descending order. (a) presents predicted weight values trained with different Dirichlet prior hyper-parameters $\alpha$. ``w/o'' means no KL loss. In (b) we plot the variance (green shading) of weight values for Dirichlet prior hyper-parameter $\alpha=1$. The above results are collected during inference on a randomly selected video clip from VFHQ-Test dataset for blind face restoration.} 
    \vspace{-2mm}
    \label{fig:visualization_weight}
\end{figure}

\paragraph{Dirichlet Prior.} We investigate the effects of Dirichlet prior hyper-parameters $\alpha$ for blind video face restoration on VFHQ-Test dataset. The results are shown in Table~\ref{tab:dirichlet_coefficient_comparison}. We found that the prior constraint by KL loss is beneficial for the face restoration task. Specifically, we achieve the lowest FVD when the $\alpha$ takes 1.0. In Figure~\ref{fig:visualization_weight}, we visualize the weight vector $\hat{\mathbf{w}}$ predicted on an example image. In Figure~\ref{fig:visualization_weight} (a), for each ``pixel" we sort the weight for code items in descending order, and compute the average weight vector over all the pixels. We found when $\alpha$ takes smaller value, the sampled weights focus on the first fewer items. When $\alpha$ takes larger value, the sampled weights are more uniform. We also show the variance of the weight values when $\alpha=1$ in Figure~\ref{fig:visualization_weight} (b). The weight values are stable over different pixels.

\begin{table}[htbp]
    \centering
    \begin{tabular}{@{}c|ccccccc@{}}
        \toprule
        $\alpha$ & PSNR↑    &AKD↓   & FVD↓   & TLME↓     \\ \midrule
        w/o      & 28.929   & \textbf{2.088}  & 336.076  & \textbf{1.091}     \\
        0.0001   & 29.091   & 2.093  & 337.365  &  1.112     \\
        1.0      & \textbf{29.099}   &  2.093    &  \textbf{336.015}  &  1.107      \\ 
        10.0     & 29.003  &     2.105   &   344.922   & 1.106   \\
        \bottomrule
    \end{tabular}
    \vspace{-3mm}
    \caption{Ablation for Dirichlet prior hyper-parameter $\alpha$ for Blind Face Restoration on VFHQ-Test dataset. ``w/o'' means no KL loss.} 
    \label{tab:dirichlet_coefficient_comparison} 
    \vspace{-5mm}
\end{table}


\paragraph{Code Aggregation Strategy.} 
We evaluate three code aggregation approaches: (1) ``Top-K'' selection with hard weight assignment during training and soft averaging during inference (K=4, 16), 
(2) ``Average'' using learned soft weights without sparsity constraints, 
and (3) our probabilistic aggregation with Dirichlet prior. Table~\ref{tab:smoothing_comparison} demonstrates that our method achieves superior performance across all metrics. 
Compared to top-16, our approach improves PSNR by 0.594 dB and reduces FVD by 4.1\%, validating that explicit modeling of code combination probabilities through variational inference better captures facial feature distributions. 
The 5.7\% reduction in AKD versus the Average baseline highlights enhanced structural preservation through our sparsity-inducing Dirichlet prior.

\begin{table}[]
    \centering
    \begin{tabular}{@{}c|cccc@{}}
        \toprule
        Method & PSNR↑               & AKD↓           & FVD↓                     & TLME↓              \\ \midrule
        Top-4            &     28.178         &          2.415                &              361.160            &                1.217          \\
        Top-16            &  28.505     &   2.283  &              349.784            &    1.140                      \\
        Average  &    29.035    &   2.128        &                  342.141        &     1.114                     \\
        Ours            &   \textbf{29.099}             &     \textbf{2.093}                     &             \textbf{336.015}             &                   \textbf{1.107}       \\ 
        \bottomrule
    \end{tabular}
    \vspace{-3mm}
    \caption{Ablation study of code aggregation strategies on VFHQ-Test for blind face restoration. ``Top-K'': hard top-1 selection during training, top-K averaging during inference. ``Average'': learned soft weights without constraints.} 
    \vspace{-4mm}
    \label{tab:smoothing_comparison} 
\end{table}

\section{Conclusion}
We present a novel video restoration framework that reformulates the latent space of codebook-based VAEs as a probabilistic mixture over codebook entries via Dirichlet distributions. 
By replacing discrete quantization with continuous convex combinations regularized through variational inference, our approach enables smooth temporal transitions while preserving reconstruction fidelity. 
The spatio-temporal Transformer architecture effectively captures inter-frame dependencies, predicting Dirichlet parameters that balance the ELBO objective's reconstruction and coherence terms. 
This work bridges the gap between discrete codebook constraints and continuous video dynamics, offering a principled direction for enhancing generative models in video processing tasks.

{
    \small
    \bibliographystyle{ieeenat_fullname}
    \bibliography{main}
}


\end{document}